\documentclass[11pt,letterpaper,onecolumn]{IEEEtran}

\usepackage[pagebackref=true,breaklinks=true,letterpaper=true,colorlinks,bookmarks=false]{hyperref}
\usepackage{graphicx}
\usepackage{amsmath}
\usepackage{amssymb}

\newcommand{\R}{\mathbb{R}}

\newcommand{\ip}[3]{\left< {#1}, {#2} \right>_{#3}}

\newcommand{\comment}[1]{ }

\newtheorem{defn}{Definition}
\newtheorem{theorem}{Theorem}
\newtheorem{remark}{Remark}

\begin{document}

\title{Matching Through Features and Features Through Matching}

\author{Ganesh Sundaramoorthi and Yanchao Yang%
\IEEEcompsocitemizethanks{\IEEEcompsocthanksitem
G.~Sundaramoorthi is with the Department of Electrical Engineering and Department of Applied Mathematics and Computational Science, King Abdullah University of Science and Technology (KAUST), Thuwal, Saudi Arabia
\IEEEcompsocthanksitem
  Y.~Yang is with the Department of Electrical Engineering,
  King Abdullah University of Science and Technology (KAUST), Thuwal, Saudi Arabia 
  \protect\\
  E-mail: \{ganesh.sundaramoorthi,yanchao.yang\}@kaust.edu.sa}
}

\markboth{Technical Report}%
{Yang and Sundaramoorthi: Modeling Self Occlusions for Object Tracking}

\maketitle

\begin{abstract}
  This paper addresses how to construct features for the problem of
  image correspondence, in particular, the paper addresses how to
  construct features so as to maintain the right level of invariance
  versus discriminability. We show that without additional prior
  knowledge of the 3D scene, the right tradeoff cannot be established
  in a pre-processing step of the images as is typically done in most
  feature-based matching methods. However, given knowledge of the
  second image to match, the tradeoff between invariance and
  discriminability of features in the first image is less
  ambiguous. This suggests to setup the problem of feature extraction
  and matching as a joint estimation problem. We develop a possible
  mathematical framework, a possible computational algorithm, and we
  give example demonstration on finding correspondence on images
  related by a scene that undergoes large 3D deformation of non-planar
  objects and camera viewpoint change.
\end{abstract}

\section{Introduction}

A fundamental question of recognition from images is whether two
two-dimensional (2D) images come from the same 3D scene.  Since scenes
are three-dimensional, an intuitive approach to answer this question
is \emph{analysis by synthesis}: generate all possible scenes, and for
each scene, motions of objects in the scene, changes of ambient
illumination, test whether the scene can explain the images, and then
pick the most probable scene out of those that can explain the images; if
the probability of this scene is high enough, the images correspond.

The latter approach seems to be intractable, and therefore, there are
two existing methods in the computer vision literature to determine
whether two images correspond to the same scene. The first is the
\emph{deformable template approach} (e.g.,
\cite{grenander1996elements}). The second approach is the
\emph{feature based approach} (e.g.,
\cite{lowe2004distinctive,mikolajczyk2004scale,matas2004robust}). We
summarize the main points of both approaches next.

The deformable templates approach to determining whether image $I_1 :
\Omega \to \R$ and image $I_2 : \Omega \to \R$ ($\Omega \subset \R^2$
is the domain of the image) correspond to the same scene is to compute
the probability of transformations (arising from viewpoint /
illumination change) relating two images given $I_1$ and $I_2$. Then
it is possible to compute the maximum a-posteriori estimate of the
transformation, and its posterior probability. A sufficiently high
maximum posterior probability implies a successful match. The main
drawback of this approach is a large search space (e.g.,
transformations are an infinite dimensional space as exact scene
specific induced transformations are hard to narrow down, and a
generic class must be considered) and therefore the computational cost
is high. The technique is usually applied to problems where it is
known that the images correspond, and the transformation is needed,
e.g., optical flow \cite{horn1981determining} and medical image
registration \cite{beg2005computing}.

The key idea in feature\footnote{Note that a feature is a statistic of
  the image, i.e., $F : \mathcal{I} \to \mathcal{F}$, where $\mathcal I$ is
  the set of images and $\mathcal F$ is the feature set.}  based
approaches in computer vision for determining image correspondence
is that the \emph{entire} transformation in the deformable template
approach \emph{need not} be computed to determine whether two images
correspond to the same scene, it is sufficient to simply determine
whether a few key points determined from
pre-processing\footnote{Pre-processing implies the representation of
  the image by features. Therefore, the images $I_1$ and $I_2$ are represented by
  features $F_1=F(I_1)$ and $F_2=F(I_2)$.} each of the images
match. The advantage of this approach is obviously computational
speed, the drawback may be that there are not enough keypoints to
recover the entire transformation, and a possible increase of false
positive image matches. The key question in this approach is
\emph{what features to use?}. There is generally no agreement on this
question, with many different features designed, e.g.,
\cite{lowe2004distinctive,mikolajczyk2004scale,matas2004robust}.

\emph{What's a good feature?} The feature should at least be
\emph{insensitive to the image formation process}, the confounding
issues in the formation of images are: viewpoint, illumination,
quantization, and noise, which we call \emph{nuisances}. The features
should also maintain the essence of the image in order to be
discriminative enough to match.  Some researchers have looked into
\emph{nuisance invariant image representations}
\cite{burns1992non,chen2000search,sundaramoorthi2009set} and more
recently \cite{poggio2011,mallat2012group}, but the right level of
invariance versus discriminability seems unresolved.

In \cite{soatto2009actionable}, it is stated that no pre-processing%
\footnote{Except in the case of sufficient statistics. A sufficient
  statistic $Y$ of an image $I\in \mathcal I$ with respect to an
  underlying variable $\theta$ (i.e., given $\theta \to I \to Y$ forms
  a Markov Chain), then a sufficient statistic $Y$ satisfies
  $i(\theta; Y)=i(\theta; I)$ where $i$ denotes mutual
  information. Therefore, $Y$ loses no information about $\theta$ from
  $I$.} of the image should be done according to the Data Processing
Inequality \cite{cover2006elements}, as any estimate made from
features of the data (image) is bound to be worse than any estimate
made from the original data (image). Thus, it seems that a deformable
template approach should be used for determining whether two images
correspond to the same scene, as no pre-processed representation is
typically used. However, this seems unsatisfying as the question of
speed and a reliable algorithm arise.

In this paper, we propose an approach to combine the benefits of both
deformable templates and featured-based approaches for the problem of
whether two images correspond to the same scene in a way that does not
violate the Data Processing Inequality.  Indeed, we show that
one \emph{should not} represent the image through features, yet we
\emph{do} suggest computing features!  We show that asking the
question of \emph{how to pre-process the image to maintain the right
  level of discriminability versus invariance is {\bf not the question
    to ask}}, yet we \emph{do} suggest calculating invariants in such
a way to maintain discriminability! We derive our approach by asking
how to compute a feature of the image that maintains the right
tradeoff between invariance and discriminability, and show that 
leads to a conundrum. We then suggest the path forward, and then
derive a possible mathematical framework to do image correspondence,
combining the benefits of deformable templates and feature-based
approaches. We then build a computational algorithm to implement our
program for determining whether two images correspond to the same
scene. As a first step, we show our algorithm working on matching
corresponding pairs of images where the 3D scene undergoes deformation
resulting from 3D deformations of objects, and camera viewpoint
change.

\section{Matching With Invariants}
\label{sec:existing_paradigms}

For the purpose of this paper, we assume a Lambertian scene, and
constant illumination\footnote{Although the method we introduce later
  can also be applied to simple models of illumination change, we
  disregard it in the rest of the paper for ease of presentation.}.
As mentioned in the previous section, there are two methods to
determine whether two images $I_1$ and $I_2$ are from the same scene.
We discuss the second approach, i.e., feature-based methods, in more
detail.  Typically, the image is represented through \emph{invariants}
(or more generally \emph{insensitive features}). The idea is to factor
out the effects of the image induced transformation (arising from
viewpoint/illumination) from $I_1$ and $I_2$, and then match the
resulting representations directly.  In otherwords, compute features
of $I_1$ and $I_2$ that remain invariant to the induced
transformation, and match the features directly.  The idea being that
if we factor out the effects of the transformation from each of the
images in a pre-processing step, then the resulting representations
should be lower dimensional than the images themselves, and hence
establishing correspondence should be easier. While the approach has
led to much success, there is no general agreement over what features
to use, whether they should be invariant, whether they should be
insensitive, and there is no accepted framework for constructing them.

We believe the features used for establishing correspondence should be
invariant to viewpoint (and other nuisances, but let us not dwell on
that now for simplicity), but the question remains, \emph{invariant to
  what transformations} (arising from viewpoint)?  Based on early work
\cite{burns1992non}, it was concluded that general viewpoint
invariants of the \emph{geometry of the 3D object} cannot be computed
from a 2D projection on the plane (as they don't exist). This has led
to computation of invariant features restricted to the case when the
3D object is flat\footnote{These are actually photometric viewpoint
  invariants not geometric viewpoint invariants.} (e.g., the SIFT
detector \cite{lowe2004distinctive}, or affine invariants
\cite{mikolajczyk2004scale}). However, it might be the case that these
are not adequate for a non-flat world. More recently, it was shown
that away from occlusions, full viewpoint invariant features of
\emph{the photometry of the 3D object} can be computed from a single
2D image \cite{sundaramoorthi2009set} and a characterization of all
such invariants was given, i.e., the \emph{maximal invariant}. Indeed,
it was shown that knowing nothing other than smoothness of the 3D
scene, viewpoint invariants of the 3D object's photometry \emph{exist}
and are only a sparse\footnote{Sparse in this context means a discrete
  subset of the image, and the image is considered in the continuum
  (infinite resolution).} subset of the image. The computation was
based on computing invariants to the full group of
diffeomorphisms\footnote{A diffeomorphism $\phi : \Omega \to \Omega$
  is a smooth invertible map whose inverse is also smooth.}  acting on
the plane (as that is the smallest group of image induced
transformations that can be chosen knowing nothing else about the 3D
scene, even though diffeomorphisms are a gigantic set of
transformations). While features assuming flat objects are not
invariant enough (or invariant to the wrong thing in case of
non-planarity), the invariant features computed in
\cite{sundaramoorthi2009set} may not be discriminative enough (i.e.,
it may be the case that the invariants of $I_1$ and $I_2$ match when
the images don't belong to the same scene since the invariants are
such small subsets\footnote{In \cite{sundaramoorthi2009set}, it is
  proven that the maximal viewpoint invariant structure knowing
  nothing (other than smoothness) about the 3D scene is the
  topological structure of the image, which is a discrete structure
  called the Attributed Reeb Tree (ART).} of the images). Therefore,
key questions remain - \emph{what is the right amount of invariance
  and discriminability?}, i.e., \emph{what is the right set of
  transformations?}, and \emph{how does one compute invariance to
  those transformations?}

\section{A Method for Features \emph{and} Matching}
In this section, we attempt to answer the questions posed at the end
of the previous section. The main point that we suggest is that the
questions above should \emph{not} be addressed \emph{a-priori} in the
pre-processing stage of the images $I_1$ and $I_2$ before matching,
those questions should be answered at \emph{match time} (when $I_1$
and $I_2$ are being matched). In otherwords, the choice of the class
of transformations to be invariant to (and hence the invariant
features to be computed so that matching can be done) should be
determined online during the matching process. We summarize our
reasons for this approach next in Section~\ref{subsec:online_features}
and then derive a possible mathematical framework in the subsequent
sections. Note that the idea of processing at match time is considered
in \cite{leeS11imavis}, but we arrive at that conclusion from a
different perspective (further differences are explained in
Section~\ref{sec:taehee_work}) - that of \emph{invariance versus
  discriminability}. 

\subsection{Why Should Features Be Computed Online?}
\label{subsec:online_features}
We start by noting the most general transformations induced in the
plane that can arise from viewpoint change and 3D deformation of
objects. A motion induced image transformation is described by a
piecewise diffeomorphism on a subset of the domain of interest,
$\Omega$. We define the class of piecewise diffeomorphisms as follows:
\begin{defn}
  A {\bf piecewise diffeomorphism} $\phi$ on $\Omega$ is defined as
  \begin{enumerate}
  \item a \emph{partitioning} of the domain $\{ R_i \}_{i=1}^{N}$ (the mapped sets)
    and $O$ the \emph{occluded set} ($R_i,O\subset \Omega$) such that
    \[
    \cup_{i=1}^{N} R_i \cup O = \Omega, \, R_i \cap R_j = \emptyset
    \, (i\neq j), \, R_i \cap O = \emptyset
    \]
    where $N\geq 1$ is the number of regions.
  \item and maps $\phi_i : R_i \to \phi_i(R_i) \subset \Omega$ such
    that $\phi_i$ is a diffeomorphism  
  \item $\phi : \Omega \backslash O \to \Omega$ is one-to-one
  \end{enumerate}
  We denote the set of all such $\phi$ that satisfy the above
  properties as $\mbox{PDiff}(\Omega)$.
\end{defn}
Any induced transformation on the image domain from the combined
effect of motions/deformations of objects and camera viewpoint change
is an element of $\mbox{PDiff}(\Omega)$. We note that the class
$\mbox{PDiff}(\Omega)$ is too general a set of transformations for
\emph{a given scene}. Indeed, if one were to consider all the image
induced transformations from motions of objects and/or camera of
\emph{the same} scene, then one would not arrive at the \emph{entire}
class of transformations $\mbox{PDiff}(\Omega)$. In other words,
$\mbox{PDiff}(\Omega)$ is too generic for a specific scene and perhaps
even a class of objects (e.g., all chairs). However, if one were to
consider all possible motions of all possible scenes, such
transformations would generate $\mbox{PDiff}(\Omega)$.

Let us attempt to answer a question at the end of
Section~\ref{sec:existing_paradigms}: what is the subset of
transformations of $\mbox{PDiff}(\Omega)$ to choose so that one can
compute an invariant feature to be able to match?  We will see that in
trying to answer this question, we run into a conundrum. One could try
to be invariant to the whole class $\mbox{PDiff}(\Omega)$, and match
the resulting representations. As discussed in the previous section,
in the specialized case of the subset $\mbox{Diff}(\Omega)$ of
diffeomorphisms, the maximal invariant representation is a sparse set
of the image, and it appears that such an invariant may not be
discriminative enough to match images. Hence, features that are fully
invariant to $\mbox{PDiff}(\Omega)$ would also not be discriminative
enough to match. Therefore, one needs to consider a smaller subset of
transformations that are specific to a scene.  Clearly, the feature
$F_1$ of $I_1$ should be invariant to image induced transformations
arising from \emph{the} scene that $I_1$ is generated from. Similarly,
the feature $F_2$ of $I_2$ should be invariant to transformations
arising from \emph{the} scene that $I_2$ arises from. Note that this
requires that one knows the scenes!

Since we do not have knowledge of the scene, it seems that the feature
cannot be invariant to just scene specific transformations (if we
don't know the scene, transformations are not known, and then it seems
impossible to compute features invariant to unknown transformations)!
However, suppose for the moment that $I_1$ and $I_2$ match
(i.e., correspond to the same scene), the more general case of when it
is not known that $I_1$ and $I_2$ match will have to wait until
Section~\ref{subsec:applicable_recognition}. Then knowledge of the
transformation can be obtained, i.e., we simply establish
correspondence in the images, and the transformation can be
obtained. But one cannot establish correspondence without first
deciding what features to match, and the features depend on the
transformation (since the feature should be invariant to it)! This is
a ``chicken and egg'' problem. Therefore, we suggest that {\bf one
  cannot separate the calculation of features from the process of
  establishing correspondence}. This is in contrast to the traditional
approach in computer vision, where one first computes features $F_1$
from $I_1$ in a preprocessing step, then computes features $F_2$ from
$I_2$ (independently from $I_1$) in a preprocessing step, and then the
features $F_1$ and $F_2$ are matched to establish correspondence. We
suggest setting up feature extraction and establishing correspondence
as a \emph{joint} estimation problem, which is a approach to solve
``chicken and egg'' problems \cite{shah1985boundary}.

\subsection{Energy for Joint Features and Matching}
We now illustrate one possible mathematical framework to illustrate
the idea of joint feature extraction and establishing
correspondence. We first note the following theorem regarding
simplification of $\mbox{PDiff}(\Omega)$.
\begin{theorem}[Approximation of $\mbox{PDiff}$ with $\mathbb{A}(2)$]
  Suppose that $\phi \in \mbox{PDiff}(\Omega)$ and $\varepsilon > 0$,
  then there exists $\{P_j\}$, a sub-partition\footnote{A
    sub-partition $\{P_j\}$ of $\{R_i\}$
    is such that for each $j$, there is an $i$ such that $P_j \subset
    R_i$ and $\cup_{j} P_j = \cup_{i} R_i$.} of a partition $\{R_i\}$, arising
  from $\phi$,  and affine
  transformations $A_i \in \mathbb A(2)$ such that $\phi$ is
  approximated up to error $\varepsilon$ in $C^1$-norm in each of the
  sets $P_i$, that is
  \[
  \|\phi-A_i \|_{C^1} =  \sup_{x\in P_i} |\phi(x) - A_i(x)|_2 + 
  | D\phi(x) - D A_i(x) |_2 < \varepsilon,
  \]
  where $D$ denotes the differential, and the norms on the right hand
  side are Euclidean norms.
\end{theorem}
The theorem is proved by noting that for a point $x\in
\mbox{cl}(R_i)$, the closure of $R_i$, a diffeomorphism can be
approximated within a small ball $B_{\varepsilon}(x) \cap R_i$ about
$x$ by an affine map. Since $\mbox{cl}(R_i)$ is compact, there is a
finite number of sets $B_{\varepsilon}(x_j) \cap R_i$ that cover
$R_i$. These sets form $P_j$.

\begin{remark}
  The approximation of $\mbox{PDiff}$ with affine maps is just one
  possible way to simplify $\mbox{PDiff}$ in such a way to create a
  joint problem in our framework. Better handling of perspective
  effects could be done with homographies, and the 
  algorithm we derive in Section~\ref{subsec:optimization_algorithm}
  can certainly be generalized to handle this case.
\end{remark}

Hence it is clear that we may replace any motion induced
transformation with a piecewise affine map. The partition is obviously
much finer than that of the piecewise diffeomorphism.  Without loss of
generality, we may assume that each $P_i$ of the partition is a square
patch. If this is not the case, one can break the partition into a
finer partition that is square.

The previous theorem allows us to setup a natural joint
problem. Firstly, if we are given a patch in the domain of $I_1$ in
which the transformation is known to be affine, then we know what
feature to compute - it will be the affine invariant representation of
the image restricted to the patch, the invariant can then be matched
to $I_2$ to establish correspondence. However, we do not know whether
a patch's motion is described by an affine model directly from one
image, $I_1$, \emph{itself}! We do however, have the second image,
$I_2$, to be matched, and we can certainly test a hypothesis as to
whether a given patch $P_i$'s motion is affine. If the test is
successful, we know how to compute the invariant and the
transformation, otherwise another hypothesis must be generated. This
is the main idea of the algorithm that we propose. Indeed, to
establish correspondence we determine a partitioning of the domain
into patches in which each of the patch's motion is described by an
affine motion, and this is setup as an optimization problem in the
partition $\{P_i\}$ and the affine motions $\{ A_i\}$ describing
motions of the patches\footnote{This sounds similar to a motion
  segmentation problem! An issue in motion segmentation is \emph{choosing the
    right shape of patch to match} \cite{brox2009large}, or in other formulations of
  motion segmentation, an issue is \emph{how many regions to match}
  \cite{cremers2005motion}, 
  and these practical issues have not been addressed
  to satisfaction. Previous works in
  motion segmentation have made the decision of which patches to match
  and/or how many as a pre-processing step for each image independently
  \cite{brox2009large,cremers2005motion}. This is in
  contrast to our approach in which we suggest that those decisions
  be made at match time.}.

Since there are many partitions that would satisfy the condition of
piecewise affine, we list the criteria for the partition that we would
like: 
\begin{enumerate}
\item the partition should be such that each patch fits an affine
  model for it's motion,
\item each patch should be as large as possible to have an efficient
  decomposition,
\item for each patch $P_i$, $I_1|P_i$, the image restricted to the
  patch, should be discriminative enough to establish unique
  correspondence, and
\item the partition should cover as much of $\Omega$ as possible.
\end{enumerate}

These criteria can be integrated into a joint estimation problem of
the partition $P_i$, and the affine maps $A_i$:
\begin{equation} \label{eq:energy}
  E( \{ A_i, P_i \}_{i=1}^N ) =  \sum_i d(I_1\circ A_i|P_i, I_2|P_i) + 
  \mathcal{C}(I_1|P_i,I_2) - \mbox{Area}(P_i)
\end{equation}
where the partition must satisfy the following properties:
\begin{enumerate}
\item $P_i \cap P_j = \emptyset$ if $i\neq j$
\item $\cup P_i \subset \Omega$ (the occlusion is then  $O = \Omega
  \backslash \cup_i P_i$ ). 
\end{enumerate}
The function $d : \mathcal{I} \times \mathcal{I} \to \R$ is some
similarity function defined on image patches (small when the arguments
are similar, and large otherwise). The function $\mathcal{C}$ is a
measure of non-uniqueness of the patch (we won't directly define
$\mathcal C$, but we address what properties it should have in the
construction of our algorithm in the next section). The
energy should be minimized.

To clarify the connection of the optimization problem to our stated
approach of joint feature extraction and establishing correspondence,
we make a few comments. Firstly, what are the features? The features
are simply image patches. That does not imply that features are
computed \emph{a-priori} as which patch to select is not known. In
earlier discussion, we stated that the feature should be the invariant
to the transformation. If we have localized patches, the
transformation is affine. But an image patch is not invariant to
affine transformations. 

However, since affine transformations form a group, following
discussion in \cite{sundaramoorthi2009set}, the orbit space of image
patches acted on by the group of affine transformations is the
invariant space, and each orbit is the invariant representation of the
image patch. That is, the orbit of $I$ is
\begin{equation}
  [I] = \{ I \circ A \, : \, A \in \mathbb{A}(2) \},
\end{equation}
which is the invariant representation of $I\in \mathcal I$, and 
\begin{equation}
  \mathcal I / \mathbb{A}(2) = \{ [I] \, : \, I\in \mathcal I \}
\end{equation}
is the invariant space of $\mathcal I$ under the group
$\mathbb{A}(2)$. Therefore, in order to match via invariant
representations, it is necessary to define a similarity function $D :
\mathcal I / \mathbb{A}(2) \times \mathcal I / \mathbb{A}(2) \to \R$
on the orbits, i.e.,
\begin{equation}
  D([I],[J]) = \min_{ A,B \in \mathbb{A}(2) } d( I \circ A, J \circ B ).
\end{equation}
Further, if the similarity function $d$ above is affine invariant, i.e.,
$d(I_1\circ A, I_2\circ A) = d(I_1,I_2)$, then $D$ above can we
written\footnote{We caution that we are doing a continuum analysis,
  i.e., that the images are assumed to be of infinite resolution, and
  the effects of quantization are ignored. In the case where
  quantization effects are taken into account, the simplification would
  not work.} as
\begin{equation}
  D([I],[J]) = \min_{ A \in \mathbb{A}(2) } d( I \circ A, J ).
\end{equation}
One example of an affine invariant similarity function is choosing the
normalized cross correlation:
\begin{equation}
  d(I_1, I_2) = -\frac{ \ip{I_1}{I_2}{\mathbb L^2} }
                              { \|I_1\|_{\mathbb L^2} \|I_2\|_{\mathbb
                                  L^2} }
\end{equation}
where $\mathbb L^2$ denotes the usual inner product. It is thus clear
that optimizing \eqref{eq:energy} uses a similarity function on the
invariant representation of the patches.

\subsection{Optimization Algorithm}
\label{subsec:optimization_algorithm}
We now address the optimization of the energy $E$. We devise a simple
\emph{approximate} algorithm below. Let us suppose that $\Omega = [0,
2^n-1]^2$ where $n\geq 1$. Let $2^{n_{min}}, 2^{n_{max}}$ be the
minimum and maximum patch sizes (the lengths of the sides of the
squares) that are considered (assume that $n_{min}, n_{max} << n$). We
assume that
\begin{equation}
  P_i \in \mathbb{P} = \left\{ [k,k+2^{m}]\times[l,l+2^{m}] \, : \, 
    k,l \in \{ 0,\ldots, 2^{n-m}-1\} , m\in [n_{min}, n_{max}] \right\},
\end{equation}
where $\mathbb{P}$ is the collection of all possible patches that can
be considered.  That is, the elements of the partition may be any
block of size $2^{n_{min}}$ to $2^{n_{max}}$ that is attained by
successively cutting the image into four equal blocks ($n$ times).

Suppose that we sample a subset of general linear transformations
($\mathbb{GL}(2)$ of invertible $2\times 2$ matrices; full affine is
possible, but we like to keep it simple):
\begin{equation}
  \mathbb{DGL}(2)  = \{ R_{\theta}^T \Lambda R_{\theta} : \theta \in 
  \frac{2\pi}{M_1} \times \{ 0, \ldots, M_2-1 \} , \,
  \lambda_1, \lambda_2 \in [-\lambda_{max}, \lambda_{max}],
  |\lambda_1|\geq |\lambda_2|, \, R_{\theta} \in SO(2) \}
\end{equation}
where $SO(2)$ is the group of planar rotations, $M_1, M_2\in \mathbb
N$ (natural numbers) such that $M_1\geq M_2$, $\lambda_{max}$ is the maximum
scale, and
\[
\Lambda = 
\left(
  \begin{array}{ll}
    \lambda_1 & 0 \\
    0 & \lambda_2
  \end{array}
\right), \mbox{ and } \lambda=(\lambda_1, \lambda_2).
\]

\begin{remark}
  In the above, we allow partial invariance, indeed, by choosing
  $M_1>M_2$ and $\lambda_{max} < \infty$, we do not get full
  invariance to scale and rotation. The desired level of invariance
  can be chosen, one of the options being full invariance.
\end{remark}

It will be useful in explaining our algorithm to define a
\emph{response function}: 
\begin{defn}
  Let $I_1, I_2 : \Omega \to \R^+$ two images. Let $P \in \mathbb{P}$
  be a patch in the partition. The {\bf response function} of patch $P$ of
  $I_1$ to $I_2$ localized to $N \subset \Omega$ is
  \[
  \mathcal{R}( \lambda, \theta, x | (I_1|P, I_2|N) ) = 
  \frac{
    \left[ ( I_1 \circ a_{\theta, \lambda} |P) \ast I_2|N \right](x)
  }{
    \| I_1 \circ a_{\theta, \lambda}|P \| 
    \sqrt{  (\mathbf{1}_P\ast I_2^2|N)(x) }
  }
  \]
  where $a_{\theta, \lambda} \in \mathbb{DGL}(2)$ (the origin in the
  computation of $a_{\theta, \lambda} \in \mathbb{DGL}(2)$ is the
  centroid of $P$), $\ast$ denotes convolution, and $\mathbf{1}_P$
  denotes the indicator function on $P$. When it is understood the
  patch $P$ and localized neighborhood $N$, we will suppress those
  arguments and simply write $\mathcal{R}(\lambda, \theta, x)$ or
  $\mathcal{R}(y)$ where $y=(\lambda, \theta, x)$.
\end{defn}

Our algorithm to optimize $E$ is described below. We first note that
$N_P$ will denote a localized sub-block of $\Omega$ that is centered
about the centroid of the patch $P\in \mathbb{P}$.
\begin{enumerate}
\item Let $Q$ denote a queue, initially empty. Add all patches that have
  size $2^{n_{max}}$ into $Q$.
\item Remove the head of the queue $Q$, and suppose it is $P$. For the patch
  $P$, compute the response function $\mathcal{R}(\cdot|(I_1|P, I_2|N_P))$
\item Determine the local maxima of
  $\mathcal{R}(\cdot|(I_1|P, I_2|N_P))$ in the three variables $\lambda,
  \theta, x$ : call the local maxima for the particular patch $P$: $y_1,
  \cdots, y_m$, and suppose that $\mathcal{R}(y_1)\geq \cdots \geq
  \mathcal{R}(y_m)$.
  \begin{itemize}
  \item If $\mathcal{R}(y_1) < T_1$, then $P$ has not
    matched. Subdivide $P$ into its four sub-blocks and add these
    sub-blocks to the queue $Q$ (if the sub-blocks do not go below the
    minimum patch size $2^{n_{min}}$).
  \item If $\mathcal{R}(y_1) > T_1$ and $\mathcal{R}(y_1) /
    \mathcal{R}(y_2) > T_2$, then the patch $P$ has matched, and do not
    sub-divide $P$ anymore.
  \item If $\mathcal{R}(y_1) > T_1$ and $\mathcal{R}(y_1) /
    \mathcal{R}(y_2) < T_2$, then the patch $P$ is not discriminative
    enough to match, and stop any further partitioning of $P$.
  \end{itemize}
\item Go to Step 2 if $Q$ is not empty.
\end{enumerate}
Note that $T_1>0$ and $T_2>0$ are decision thresholds and are related to
the weights on the area and patch complexity $\mathcal C$ in the
energy $E$, respectively. The result of the algorithm is a
partitioning of $\Omega$, $\{P_i\}$ (of possibly different sizes), and
the affine transformations $\{A_i\} = \{ \lambda_i, \theta_i, x_i \}$
and so the corresponding patches in $I_2$ are established. Note that it
is not necessarily true that $\cup_i P_i$ is all of $\Omega$. The
occluded set $O$ and the patches that are not discriminative enough to
match form the set $\Omega \backslash \cup_i P_i$.

One can think of our algorithm as a greedy algorithm to optimize the
underlying energy $E$. Obviously by starting with the largest possible
patches as candidate patches, we are maximizing the area of the
patches used, the fact that only patches that have $\mathcal{R} > T_1$
are accepted means that only patches with a sufficiently good match
(i.e., low $d$ in the energy) are accepted, and the fact that only
matches that are sufficiently unique are accepted implies that the
patch discriminability function $\mathcal C$ is optimized. This last
statement implies the choice of $\mathcal C$ (which was not described
earlier), that is, $\mathcal C$ is chosen large when unique correspondence
cannot be established, and small when unique correspondence is
established.  Thus, it is clear that the algorithm is a rough greedy
search to optimize the energy $E$.

\begin{remark}
  The algorithm above lends itself well to a parallel
  implementation. Indeed, all patches on the queue $Q$ can be
  processed on separate processors at once since the result of any
  patch on the queue at a current time does not impact any other
  patches on the queue at the same time.
\end{remark}

\subsection{Does the Framework Apply to Recognition?}
\label{subsec:applicable_recognition}
The framework for joint feature extraction and establishing
correspondence that was derived in the previous sections was based on
the assumption that $I_1$ and $I_2$ correspond to the same scene. It
seems that the framework may only be limited to the case of matching
images under the same scene, which has wide applicability, but seems
to be limiting if it cannot be applied to recognition. Indeed, the
framework does apply to recognition.  For example, let us consider a
simple example of the recognition problem where we have training
templates $t_1, \ldots, t_m : \Omega \to \R$ of images of objects
$o_1, o_2, \ldots, o_m$. Each template represents a different object
that can be recognized. The goal now is, given a test image $I$, to
determine which object $o_i$ the image represents. To do this, one
makes a hypothesis that $I$ and $t_1$ correspond (that is, that $I$ is
recognized as $o_1$), and the hypothesis is tested by trying to
establish correspondence as we have described in the previous
sections. If the energy $E$ for the optimal correspondence and
partition established is such that $E < T_3$ where $T_3$ is the
acceptance threshold, then $I$ and $t_1$ correspond and $I$ is
recognized as $o_1$. If $E > T_3$ then $I$ is not recognized as $t_1$,
and one proceeds to $t_2$, and test the hypothesis that $I$ is
recognized as $o_2$, and so on.

Therefore, it is clear that our joint matching and feature extraction
framework is generic enough to be applied to the problem of
recognition.

\begin{figure} 
  \centering
  \includegraphics[totalheight=2.5in]{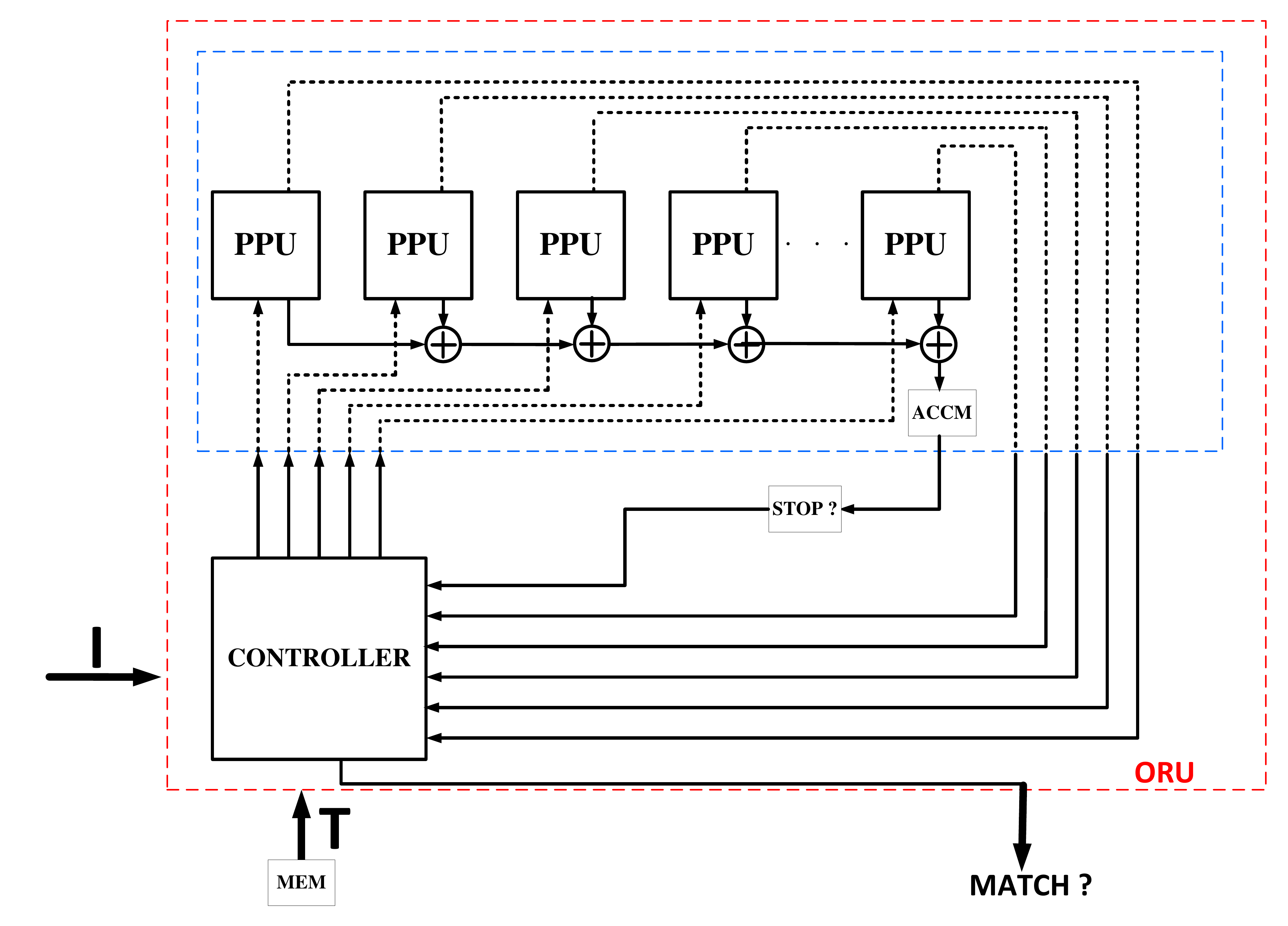}
  \caption{This is a circuit diagram for our computational algorithm
in Section~\ref{subsec:optimization_algorithm}. Such a system contains a cluster of simple primitive
units - Patch Processing Units (PPUs), which take in $I|P_i$ and
$t|N_i$ and return whether there was a match (if so, its contribution to
the energy) or not, and whether to continue processing $P_i$ by
splitting.  Notice that once the accumulation of the energies emitted
from the PPU's (collected by the ACCM) is less than the threshold,
processing is stopped (even though all patches are not processed) if
it is only desired to know whether $I$ and $t$ correspond to the
same scene. The controller manages the patches to process with the
limited available number of PPU's.  Several of the ORU's can be
connected in parallel with the test image $I$ and the training
templates $t_1,\ldots, t_m$ accessed from memory for the primitive system
discussed in Section~\ref{subsec:applicable_recognition}. }
  \label{fig:ORU}
\end{figure}

\begin{remark}
  Note that in recognition, it is only desired to determine whether
  $I$ corresponds to $t_i$. In our formulation in the
  previous section, once a patch is successfully matched, its
  contribution to the energy is known, and its contribution will not
  be changed as further patches are processed. Therefore, once the
  accumulated energy of the patches matched go below the threshold
  $T_3$, a successful match can be reported and all further processing
  of patches in the queue $Q$ can be stopped. A circuit diagram of
  this idea is illustrated in Figure~\ref{fig:ORU}.
\end{remark}

\begin{remark}
  Several of the ORU's can be connected in parallel and attached to
  the test image $I$. To construct a primitive object recognition
  system to recognize a person (assuming Lambertian scene, constant
  illumination, no noise), one could take as training data few
  snapshots of the head of the person (e.g., frontal view, back view,
  and a sideways view), and each of these templates could be attached
  to a separate ORU. If the test image at a new pose $I$ is inputted,
  at least one of the ORU's would indicate a match, thus constructing
  a very primitive object recognition system.
\end{remark}

\section{Discussion}

\subsection{Relation to 3D Reconstruction (Static Case)}
One can think of our approach (in the special case of a static scene)
as related to \emph{matching by attempting 3D reconstruction}. The
results of the structure from motion problem
\cite{yezziS03cvpr,hartley2000multiple} (i.e., in the case of a static
scene not a dynamic one) indicate that (with suitable priors), the 3D
geometry of the scene can be recovered, so the 3D scene is
\emph{essentially coded in two images of the scene}\footnote{To be
  precise, the part of the 3D scene that is co-visible in the two
  images is encoded in the two images, up to priors.}.  Therefore, once
the correct training template and the image to be recognized are
associated, the 3D object is known (the part that is
co-visible). Hence, it is clear that our mathematical framework and
proposed procedure for recognition is equivalent to attempting 3D
reconstruction, and determining whether the reconstruction has been
successful (through threshold $T_3$), but reconstruction is not
actually done. The 3D prior on scene is equivalent to a prior on the
image-induced transformation, the prior is that coarse transformations
are favored over finer transformations as our algorithm is a coarse to
fine approach. This translates to a 3D prior that the scene is as flat
whenever possible, but since fine transformations (i.e., small patch
sizes) can also be used, the prior used is not a flat scene globally.

\subsection{Relation to Existing Work}
\label{sec:taehee_work}
In \cite{leeS11imavis}, it is stated that nuisances must be eliminated
at decision time (a point that is the basis of our approach), however,
the reason stated in \cite{leeS11imavis} for nuisances being
eliminated at decision time is \emph{far different} than our stated
reasons of the \emph{tradeoff between invariance and
  discriminability}. The reason given in \cite{leeS11imavis} is
``... for a nuisance, to be eliminated in pre-processing without loss
of discriminative power, it should be invertible and commutative
[w.r.t. non-invertible nuisances, i.e., noise, quantization, or
occlusion],'' suggesting that pre-processing should be avoided because
of the non-invertible nuisances.  Our results state that \emph{no
pre-processing should be done to eliminate nuisances}. For the case of
viewpoint, the effects of viewpoint cannot be removed in a
pre-processing step without inevitably losing discriminability (even
without interaction by non-invertible nuisances); eliminating
viewpoint may only done with knowledge of the second image, not in a
pre-processing step. Further, our mathematical framework and
computational algorithm are completely different. Also, the work in
\cite{leeS11imavis} applies to tracking (small deformations), but our
algorithm is built for large deformations and motion.

\subsection{Comment on Real Physical Systems}
Our analysis and mathematical framework are done in the
\emph{continuum}, i.e., we haven't addressed the nuisances of
quantization (both in \emph{time} and \emph{space}) and
noise\footnote{We have addressed the issue of viewpoint, 3D
  deformation, and occlusion - great challenges according to
  \cite{poggio2011}.  The framework also applies to contrast change (a
  weak model of illumination).}, that are present in real systems. In
the continuum, our results state that the best approach to determine
the right level of invariance and discriminability is to determine
that when the test image (acquired at infinite resolution) is
available and is being matched to the stored infinite dimensional
representation of the training image. One may argue that this is too
idealistic, and that the issue that one should address is how to find
the right tradeoff between how much (and what) of the test image and
training image to throw away such that the recognition system makes
the least error. In this regard, the work of \cite{leeS11imavis} is
interesting, which looks into this question. We agree with that
approach, however, it is surely important to analyze the idealistic
case as it provides the limits as to how the
discriminability/invariance issue is addressed as the computational
and memory resources are increased.

\section{Experiments}
\begin{figure}
  \centering
  \includegraphics[totalheight=2.2in]{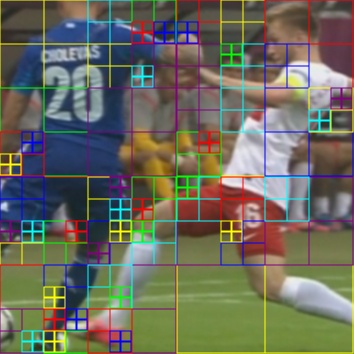}
  \includegraphics[totalheight=2.2in]{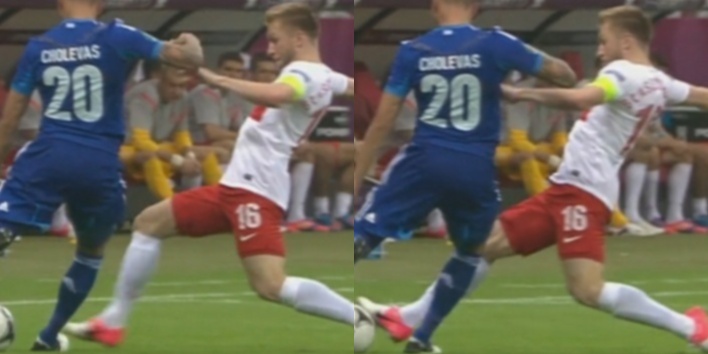}\\
  \includegraphics[totalheight=2.2in]{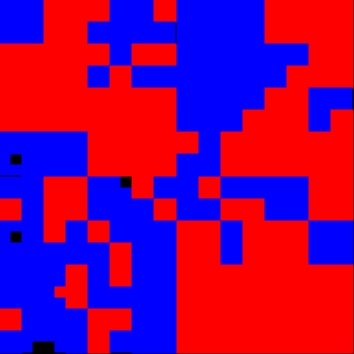}
  \includegraphics[totalheight=2.2in]{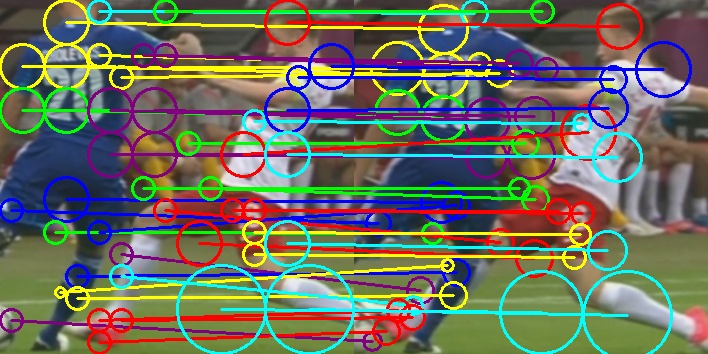}
  \caption{This figure illustrates our method working. Top left: $I_1$
    with the patches that were processed in our algorithm (i.e.,
    either matched or rejected due to being not discriminative enough)
    superimposed. Top right: $I_1$ and $I_2$.  Bottom left: the
    decisions that were made in the algorithm (i.e., red-matched, blue
    - rejected for non-discriminability, black - rejected for being
    too small). Bottom right: matching patches. It is interesting to
    see that most processing is happening near occlusions, as
    expected.}
  \label{fig:method_illustration}
\end{figure}

In this section, we demonstrate our algorithm working on establishing
correspondence of two images from the same scene. We are interested in
establishing correspondence of images taken from \emph{dynamic} 3D
scenes with camera viewpoint change, 3D object deformations,
occlusions, and \emph{close up shots} (so that there are significant
changes of shape in the 2D images, and the time between frames is
large so trackers and optical flow methods would fail). To this end,
we obtain corresponding images from closeups in sports video. As there
is no methodology for testing ground truth (e.g., epipolar geometry
doesn't apply to deforming objects), we verify matches manually (and
this cannot be done on large-scale
unfortunately). Figure~\ref{fig:method_illustration} shows the
decisions that our algorithm has made to establish matches, and the
final matching on a sample image. Note for all experiments, we choose
$\lambda_1=\lambda_2$ (i.e., only uniform scalings) for simplicity and
speed.

Although our main contribution is a methodology for how to design a
feature exhibiting the right tradeoff between invariance and
discriminability, which leads to a joint feature extraction and
matching framework, we show comparison to feature-based matching
methods. We are not trying to prove that we out-perform every method,
but simply try to give some idea how our algorithm compares to methods
that do not setup the problem jointly. In all the experiments, we have
spent much time tuning the parameters of SIFT (indeed, we have run
SIFT on many different parameter configurations using the VLFeat
\cite{vedaldi2010vlfeat} 
implementation, and show the best results). We also compare to
Harris-Affine with a SIFT descriptor, and online code was used that
did not allow for changing parameters \cite{mikolajczyk2005performance} (we
tried many different combinations of detectors and descriptors, but
the combination of Harris-Affine and a SIFT descriptor, and the SIFT
\cite{lowe2004distinctive} were best).  Figure~\ref{fig:results} shows
the results. First, we show our algorithm working on the Graffiti
Dataset (where classical methods apply as the scene is flat). Our
method gives comparable results on these images. Note that for very
large perspective change, our current implementation of affine
transforms on patches is not sufficient (although it is straight
forward to add perspective to our model).

In the next images of Figure~\ref{fig:results}, we show only the
foreground feature matches as we want to test our algorithm on 3D
deformations and viewpoint change for non-flat scenes. Again, we have
exhaustively searched over all parameters of SIFT and displayed the
best results. It can be seen that the proposed method captures much
more of the foreground area than SIFT and Harris-Affine, and makes few
mistakes.

\begin{figure}
  \centering
  \includegraphics[totalheight=2in]{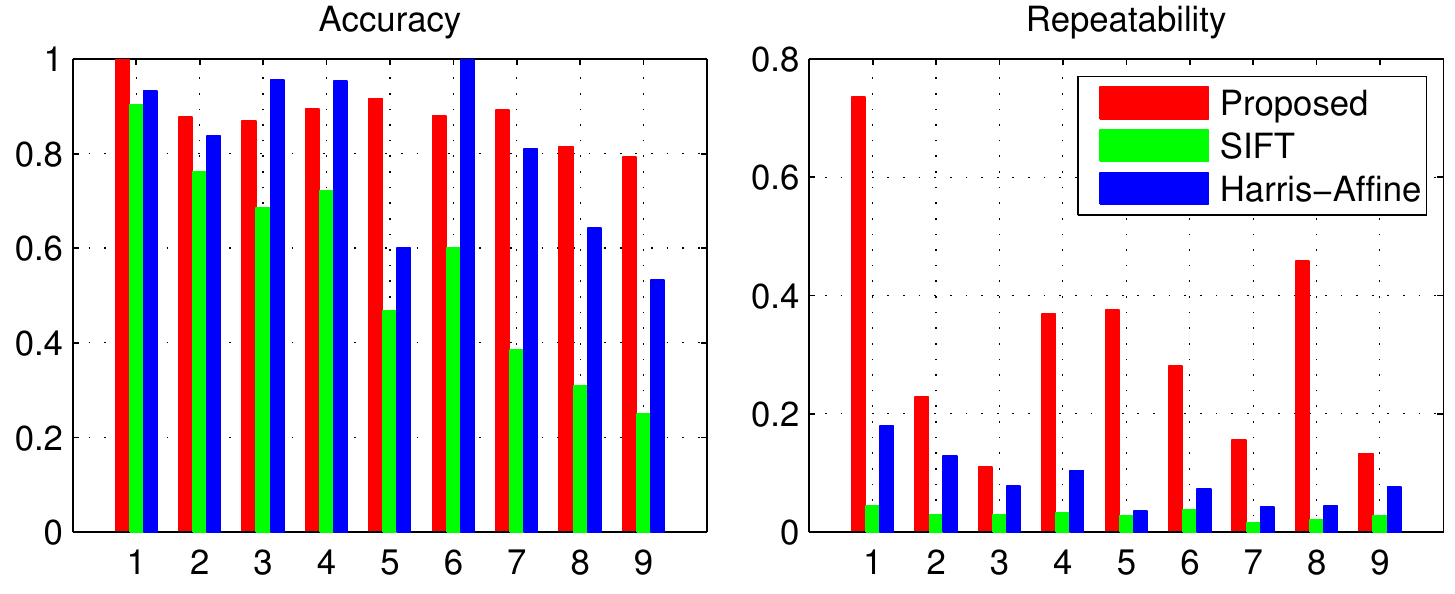}
  \caption{This figure shows quantitative comparison of the proposed
    method to standard methods, SIFT and Harris Affine (with a SIFT
    descriptor) using standard metrics - repeatability and
    accuracy. Our method yields higher repeatability, and many times
    higher accuracy (Harris affine achieves high accuracy by matching
    only a few features). The proposed method is a different paradigm
    than current feature-based matching, and so standard evaluation
    metrics may be insufficient to illustrate the power of our method.}
  \label{fig:bar_graph}
\end{figure}

A quantitative assessment is given in Figure~\ref{fig:bar_graph} using
standard evaluation metrics, accuracy and repeatability. For standard
feature matching methods, these are computed as follows:
\begin{equation}
  \mbox{repeatability} = \frac{ \mbox{\# features matched} }{ 
    \mbox{min \# of features detected in $I_1$ and $I_2$} }, \,\,
  \mbox{accuracy} = \frac{ \mbox{\# correct matches} }{ 
    \mbox{\# features matched} }.
\end{equation}
The proposed method mostly performs better. Note that in the
computation of repeatability for our method, we use the number of
patches that were considered as the denominator in the above formula,
others remain the same.  In Figure~\ref{fig:bar_graph}, the $x-$axis
in the bar graphs indicate the image number in the same ordering as
the images stacked in Figure~\ref{fig:results} on the left column.  We
wish to point out that it may be the case that the standard evaluation
metrics do not capture the full benefit of our method, as can be seen
in Figure~\ref{fig:results}: in contrast to other feature-based
matching techniques, our method captures almost the whole region of
the foreground (minus occlusion) and most matches are correct whereas
the other approaches cover very little of the foreground area.

\def\fPath{figures}

\def\fHeight{0.865in}
\begin{figure*}
  \centering
  \includegraphics[totalheight=\fHeight]{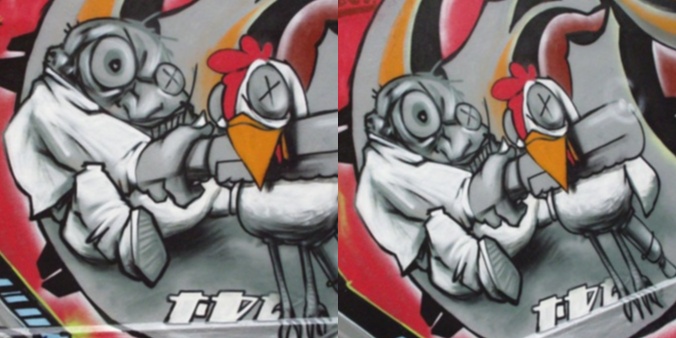}
  \includegraphics[totalheight=\fHeight]{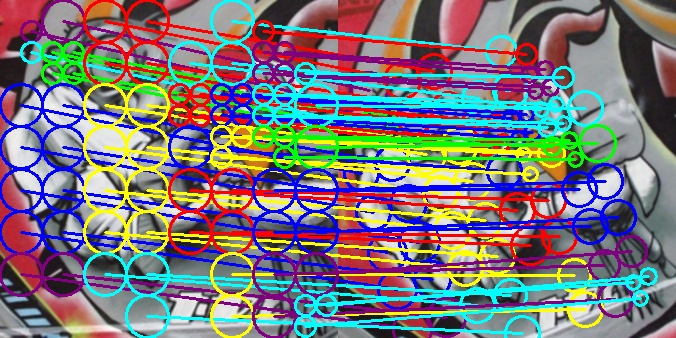}
  \includegraphics[totalheight=\fHeight]{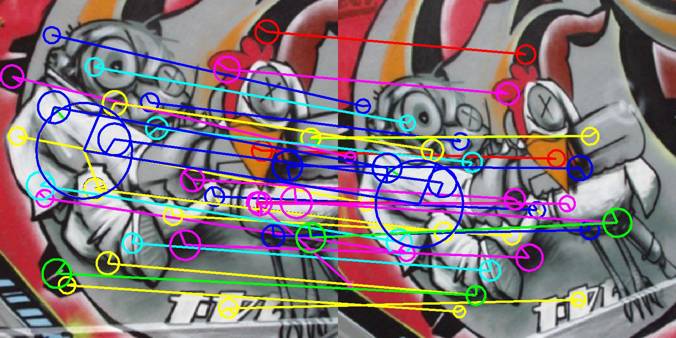}
  \includegraphics[totalheight=\fHeight]{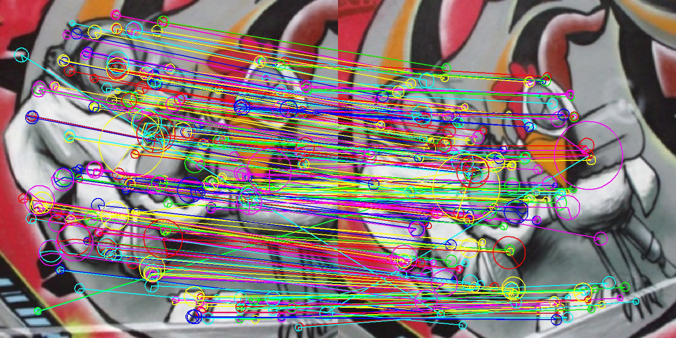}\\
  \includegraphics[totalheight=\fHeight]{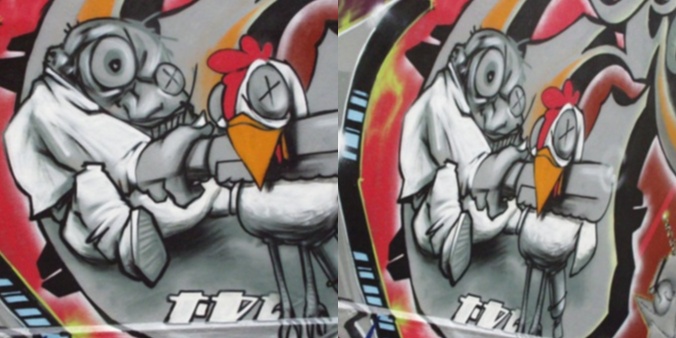}
  \includegraphics[totalheight=\fHeight]{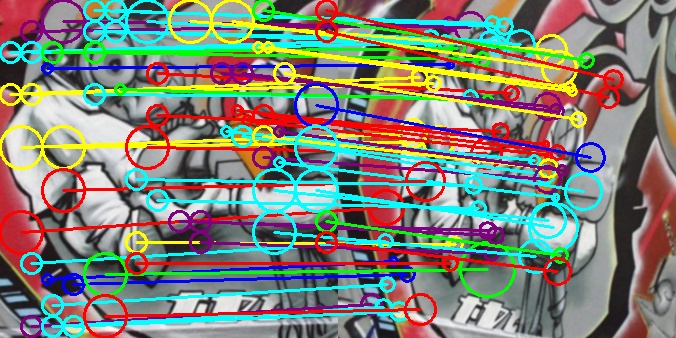}
  \includegraphics[totalheight=\fHeight]{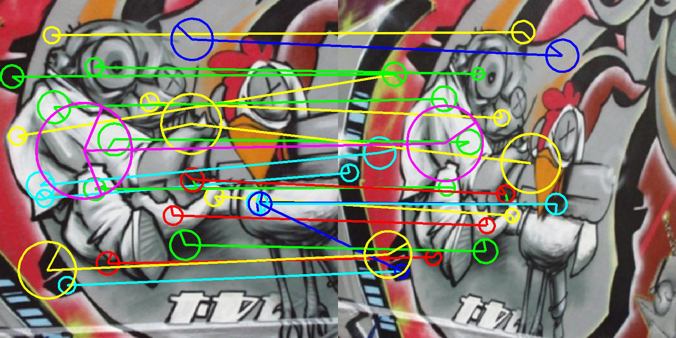}
  \includegraphics[totalheight=\fHeight]{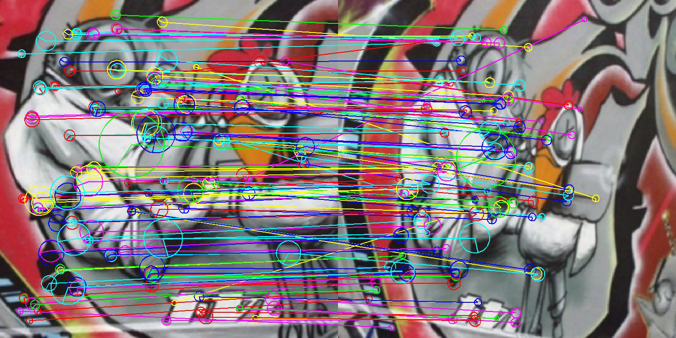}\\
  \includegraphics[totalheight=\fHeight]{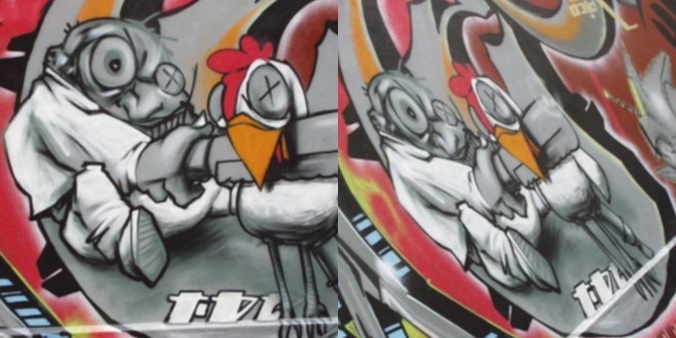}
  \includegraphics[totalheight=\fHeight]{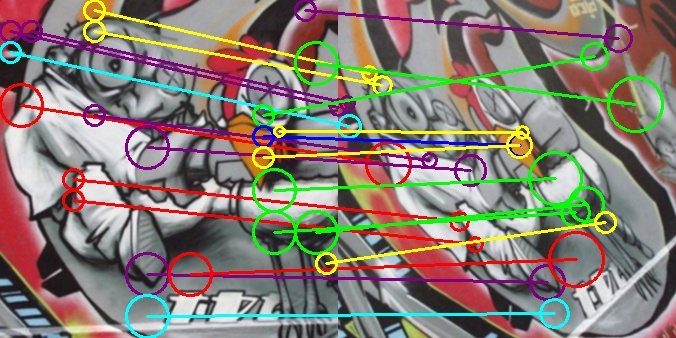}
  \includegraphics[totalheight=\fHeight]{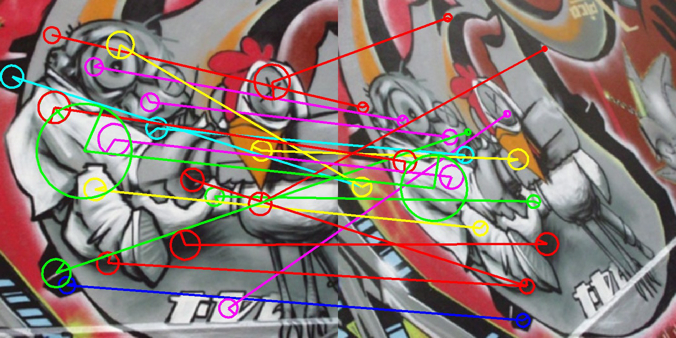}
  \includegraphics[totalheight=\fHeight]{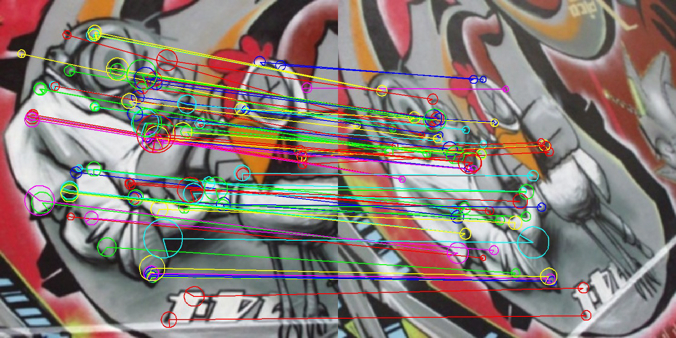}\\
  \includegraphics[totalheight=\fHeight]{\fPath/1/combined}
  \includegraphics[totalheight=\fHeight]{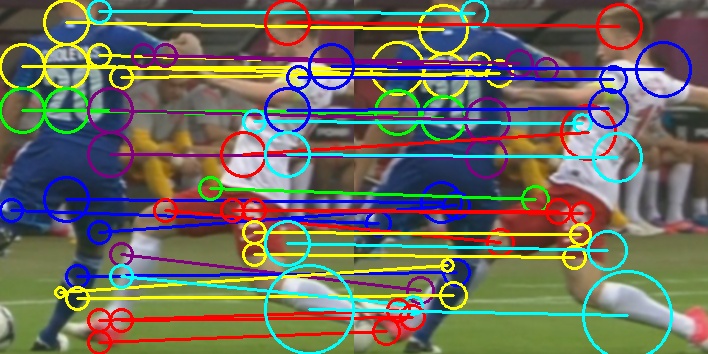}
  \includegraphics[totalheight=\fHeight]{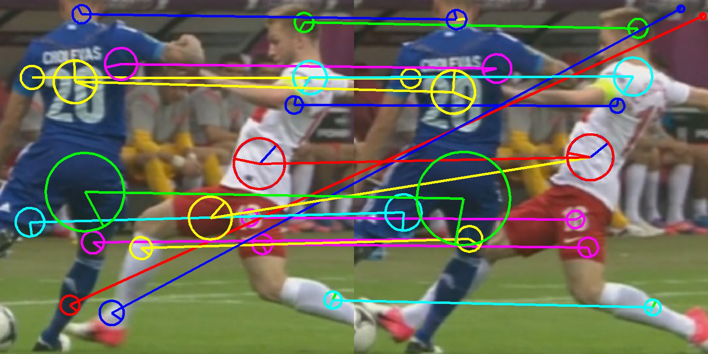}
  \includegraphics[totalheight=\fHeight]{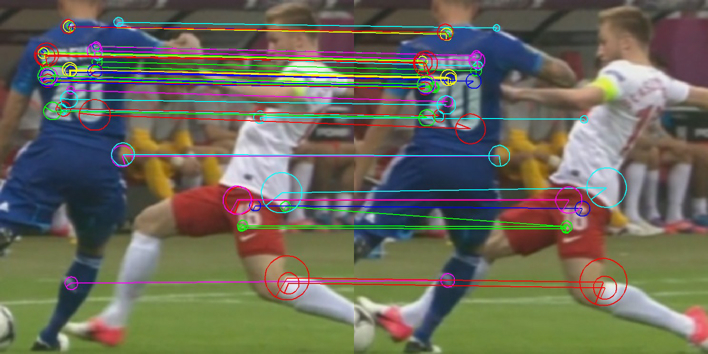}\\
  \includegraphics[totalheight=\fHeight]{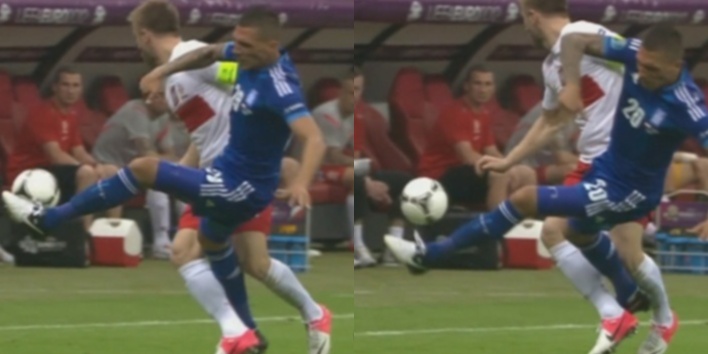}
  \includegraphics[totalheight=\fHeight]{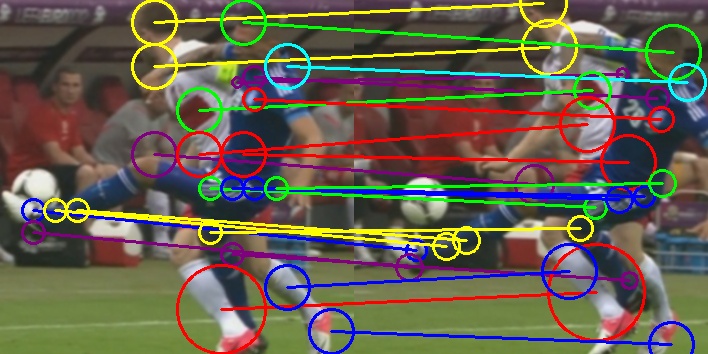}
  \includegraphics[totalheight=\fHeight]{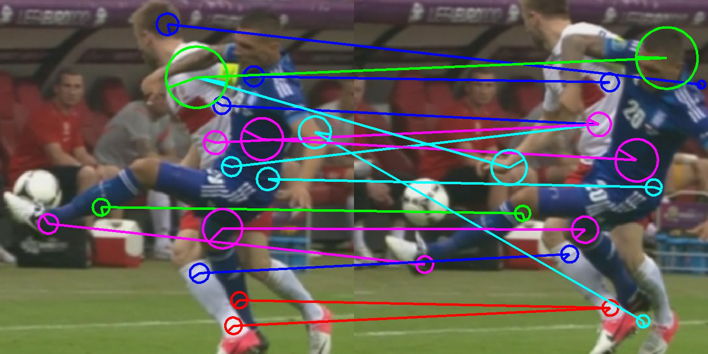}
  \includegraphics[totalheight=\fHeight]{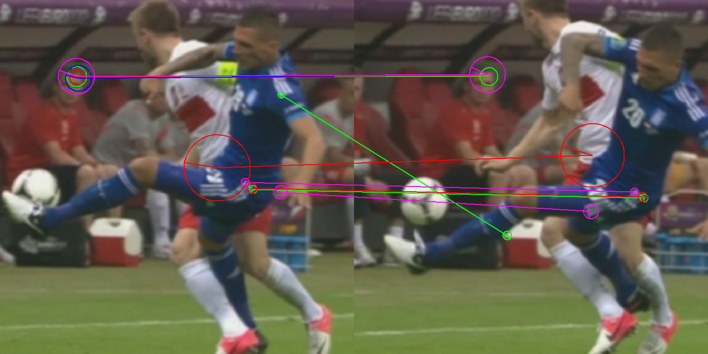}\\
  \includegraphics[totalheight=\fHeight]{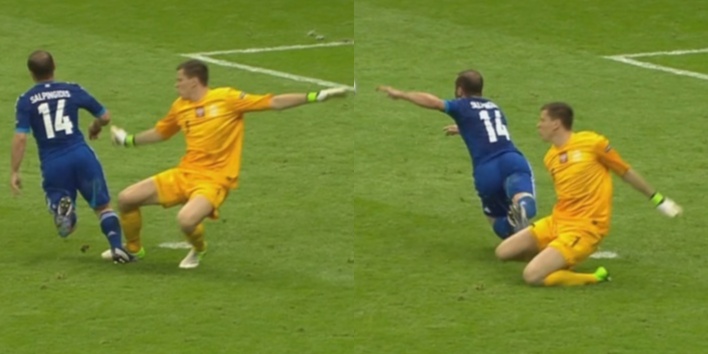}
  \includegraphics[totalheight=\fHeight]{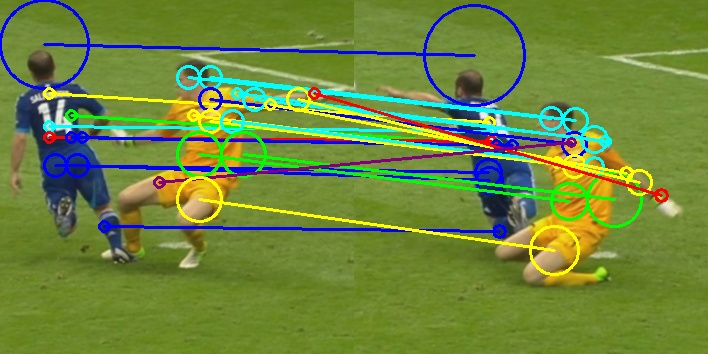}
  \includegraphics[totalheight=\fHeight]{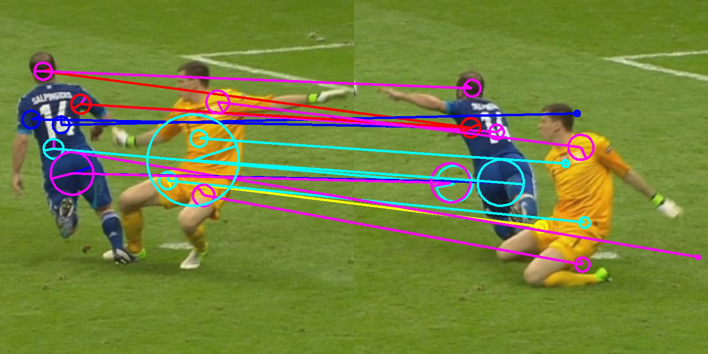}
  \includegraphics[totalheight=\fHeight]{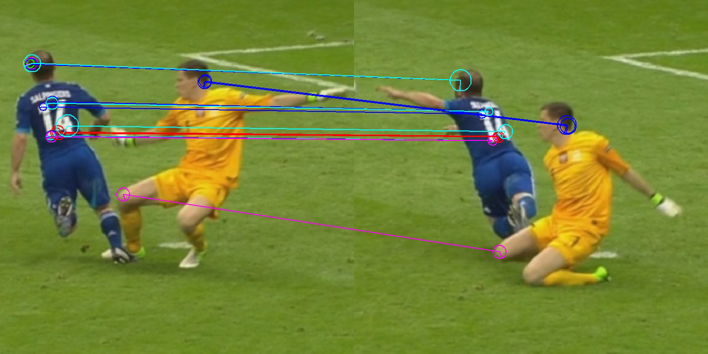}\\
  \includegraphics[totalheight=\fHeight]{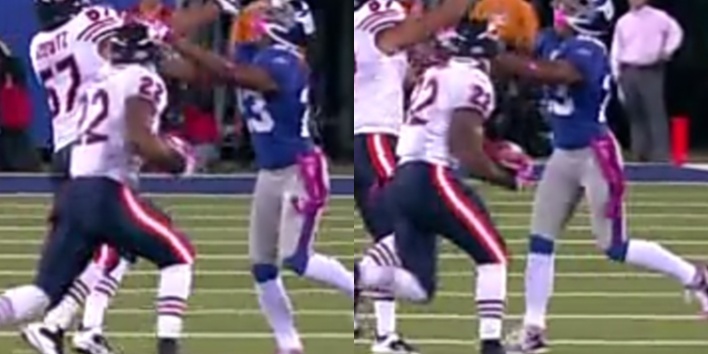}
  \includegraphics[totalheight=\fHeight]{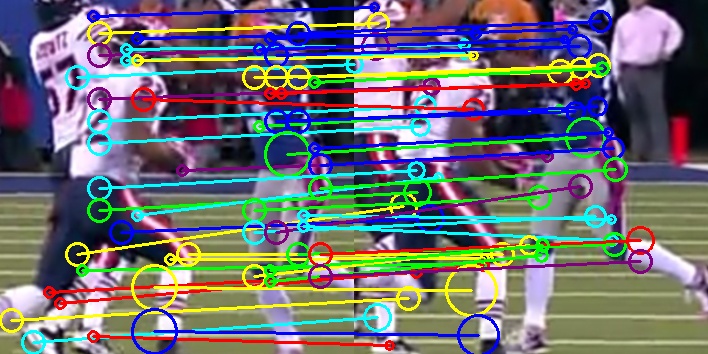}
  \includegraphics[totalheight=\fHeight]{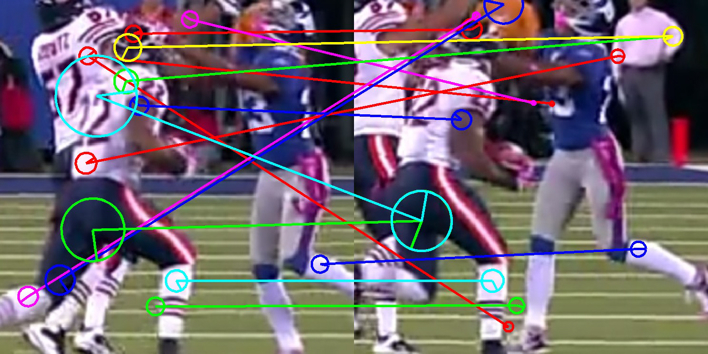}
  \includegraphics[totalheight=\fHeight]{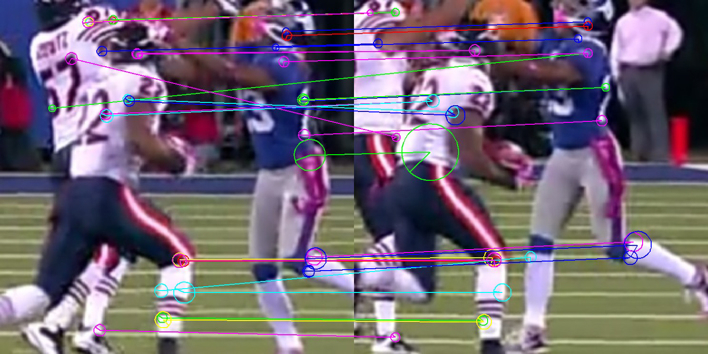}\\
  \includegraphics[totalheight=\fHeight]{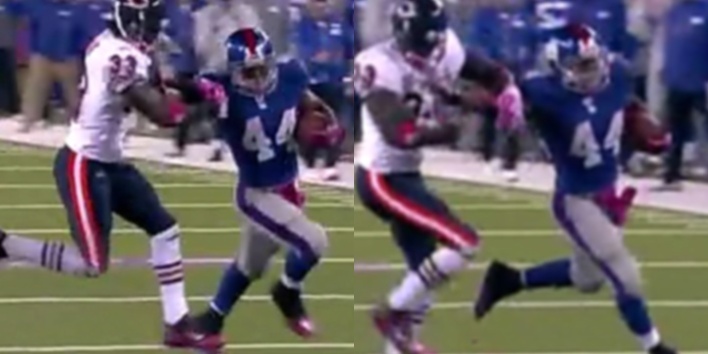}
  \includegraphics[totalheight=\fHeight]{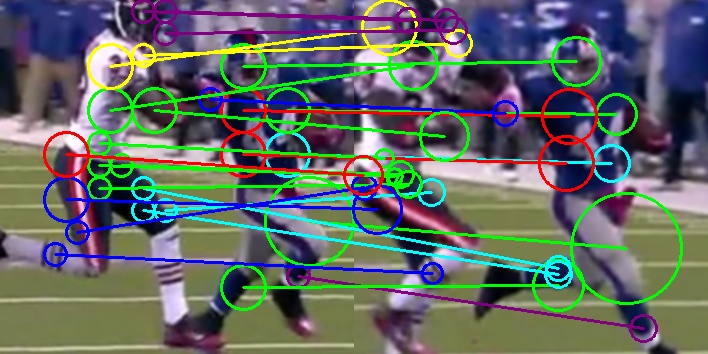}
  \includegraphics[totalheight=\fHeight]{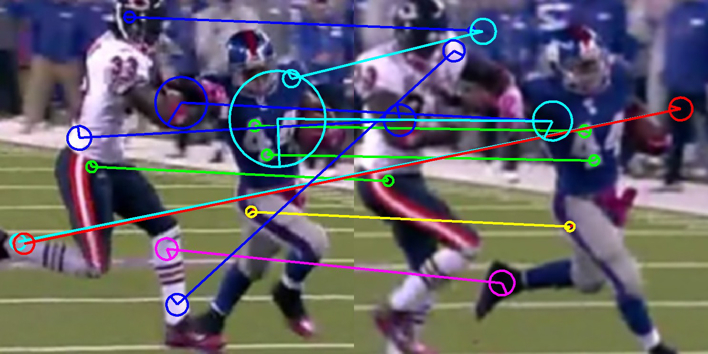}
  \includegraphics[totalheight=\fHeight]{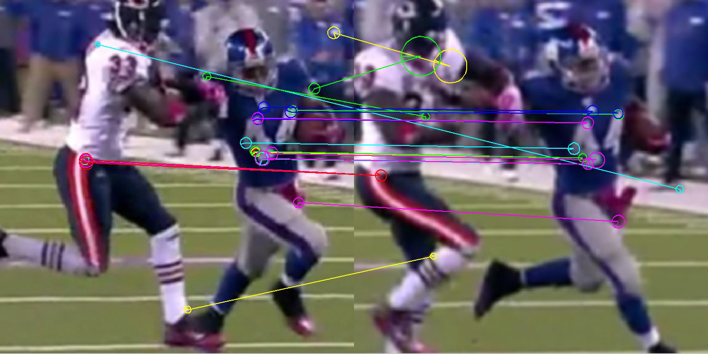}\\
  \includegraphics[totalheight=\fHeight]{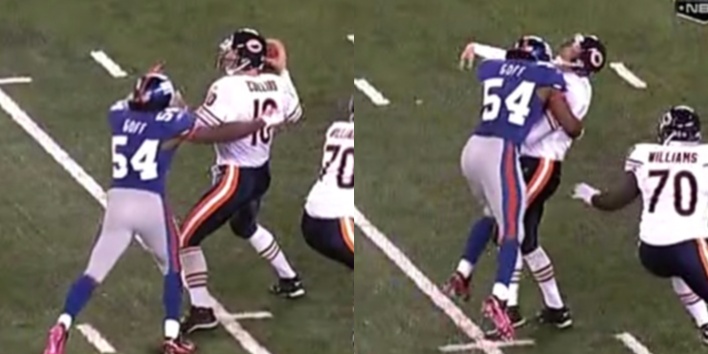}
  \includegraphics[totalheight=\fHeight]{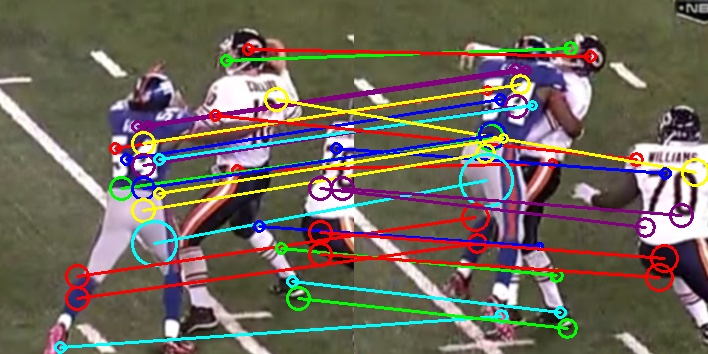}
  \includegraphics[totalheight=\fHeight]{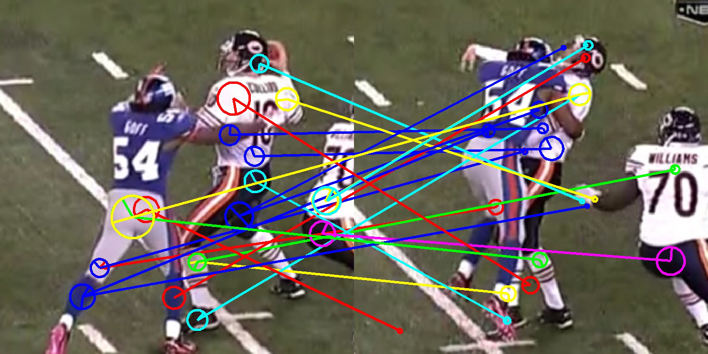}
  \includegraphics[totalheight=\fHeight]{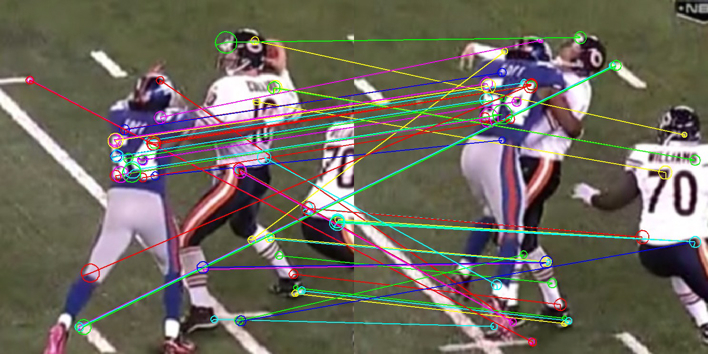}\\
\caption{Left column: $I_1$ and $I_2$, second column: results of the
  proposed method, third column: results of SIFT
  \cite{lowe2004distinctive}, and fourth column: results of
  Harris-Affine detector \cite{mikolajczyk2004scale} and a 
  SIFT descriptor.}
\label{fig:results}
\end{figure*}

\section{Conclusion}
We have addressed the question of how to construct features in a way
that has the correct tradeoff between invariance and
discriminability. We showed that the question can only be answered at
match time. In otherwords, we have shown that the right level of
invariance versus discriminability cannot be determined in a
pre-processing step (without additional information of the 3D
scene). This has led us to a joint problem of feature extraction and
matching. We have created an effective computational algorithm. Our
algorithm is designed for matching under large 3D object deformation,
camera viewpoint change, and occlusions. We have illustrated our
method on images from the same scene that exhibit the aforementioned
phenomena, and showed some comparison to standard feature based
methods where feature computation is done in a pre-processing
step. Experiments suggest that our method is more effective in the
case of large 3D deformation, and camera viewpoint for non-planar
scenes.

\section*{Acknowledgements}
GS would like to acknowledge Stefano Soatto for many discussions
regarding viewpoint invariants, from the initial work in
\cite{sundaramoorthi2009set} to the present.  GS and YY would like to
acknowledge Naeemullah Khan and Marei Garnei for help with experiments
comparing to current feature matching techniques. GS wishes to thank
Khaled Salama and Mahmoud Ouda for help making Figure~\ref{fig:ORU}.

\bibliographystyle{IEEEtran}
\bibliography{features}

\end{document}